\documentclass[sigconf,natbib=true]{acmart}

\usepackage{booktabs}
\usepackage{graphicx}
\usepackage{enumitem}
\usepackage{multirow}
\newcommand{\etal}{\textit{et al.}}
\newcommand{\partitle}[1]{\vspace{1mm}\noindent\textbf{#1}}

\usepackage{lipsum}
\usepackage{float}

\AtBeginDocument{%
  \providecommand\BibTeX{{%
    \normalfont B\kern-0.5em{\scshape i\kern-0.25em b}\kern-0.8em\TeX}}}

\copyrightyear{2022}
\acmYear{2022}
\setcopyright{acmlicensed}\acmConference[CIKM '22]{Proceedings of the 31st ACM International Conference on Information and Knowledge Management}{October 17--21, 2022}{Atlanta, GA, USA}
\acmBooktitle{Proceedings of the 31st ACM International Conference on Information and Knowledge Management (CIKM '22), October 17--21, 2022, Atlanta, GA, USA}
\acmPrice{15.00}
\acmDOI{10.1145/3511808.3557662}
\acmISBN{978-1-4503-9236-5/22/10}

\begin{document}

\title{On~the~Impact~of~Speech~Recognition~Errors in~Passage~Retrieval for~Spoken~Question~Answering }

\author{Georgios Sidiropoulos}
\affiliation{%
  \institution{University of Amsterdam}
  \city{}
  \country{}}
\email{g.sidiropoulos@uva.nl}

\author{Svitlana Vakulenko}
\authornote{Research conducted when the author was at the University of Amsterdam.}
\affiliation{%
  \institution{Amazon Alexa AI}
  \city{}
  \country{}}
\email{svvakul@amazon.com}

\author{Evangelos Kanoulas}
\affiliation{%
  \institution{University of Amsterdam}
  \city{}
  \country{}}
\email{e.kanoulas@uva.nl}

\begin{abstract}
Interacting with a speech interface to query a Question Answering (QA) system is becoming increasingly popular. Typically, QA systems rely on passage retrieval to select candidate contexts and reading comprehension to extract the final answer. While there has been some attention to improving the reading comprehension part of QA systems against errors that automatic speech recognition (ASR) models introduce, the passage retrieval part remains unexplored. However, such errors can affect the performance of passage retrieval, leading to inferior end-to-end performance. To address this gap, we augment two existing large-scale passage ranking and open domain QA datasets with synthetic ASR noise and study the robustness of lexical and dense retrievers against questions with ASR noise. Furthermore, we study the generalizability of data augmentation techniques across different domains; with each domain being a different language dialect or accent.
Finally, we create a new dataset with questions voiced by human users and use their transcriptions to show that the retrieval performance can further degrade when dealing with natural ASR noise instead of synthetic ASR noise.

\end{abstract}



\maketitle

\section{Introduction}
Nowadays users interact with a wide range of commercial Question Answering (QA) systems via speech interfaces. Millions of users are voicing their questions on virtual voice assistants such as Amazon Alexa, Apple Siri, or Google Assistant through their smart devices. Such voice assistants do not only increase the convenience with which users can query them but can support users with visual and motor impairments for which the use of conventional text entry mechanisms (keyboard) is not applicable \cite{DBLP:conf/chi/PradhanMF18}. Despite the popularity of voice assistants among users globally and the advancements in spoken-language understanding \cite{DBLP:conf/emnlp/BastianelliVSR20, DBLP:journals/corr/abs-2204-08582}, there are surprisingly limited efforts in studying spoken QA and its limitations.

The majority of research focuses on reading comprehension as a component of spoken QA \cite{DBLP:conf/interspeech/LeeWLL18, DBLP:conf/emnlp/FaisalKAA21, DBLP:conf/eacl/RavichanderDRMH21}. In detail, previous works studied the case where the provided question includes noise introduced by an automated speech recognition (ASR) system; audio is converted to text before reading comprehension is performed. 
Ravichander \etal{}~\cite{DBLP:conf/eacl/RavichanderDRMH21} showed that ASR noise not only dramatically affects the performance of transformer-based reading comprehension models but also that it is a more challenging type of noise compared to the noise generated from keyboard mistyping or faulty machine translation. 
Faisal \etal{}~\cite{DBLP:conf/emnlp/FaisalKAA21} showed that background differences in users, such as their accent, can affect the performance of reading comprehension models differently.

Even though robustifying reading comprehension against ASR noise is essential for extracting the answer to a question, as a subsequent step of passage retrieval, it is bounded by the ability of the QA system to retrieve the relevant passages. A typical QA pipeline consists of an efficient retriever that reduces the search space from millions of passages to the top-k and a reader that extracts the answer. Dense passage retrieval has become a new paradigm
to retrieve relevant passages, setting the state-of-the-art performance in several leaderboards. Inferior retrieval of the relevant passages will negatively affect the performance of the overall system.

Typically, state-of-the-art dense retrieval models are evaluated on clean datasets with noise-free questions. However, questions posed to real-world QA systems are prone to errors. Therefore, these models will encounter noisy questions when deployed in real-world applications, affecting their performance. User-generated textual questions can include typos such as keyboard typos due to fast typing, misspellings and phonetic typing errors (for words with close pronunciation). Recent works showed that even state-of-the-art dense retrieval models are not robust against simple typos \cite{DBLP:journals/corr/abs-2204-00716, DBLP:conf/sigir/SidiropoulosK22,DBLP:conf/emnlp/ZhuangZ21}. Sidiropoulos and Kanoulas \cite{DBLP:conf/sigir/SidiropoulosK22} showcased the dense retrievers' lack of robustness to typos in the question and proposed a combination of data augmentation with a contrastive loss to robustify the model. Zhuang and Zuccon \cite{DBLP:journals/corr/abs-2204-00716} increased the robustness of dense retrievers against typos by replacing the extremely sensitive to typos WordPiece tokenizer with the Character-CNN module and further combined it with a knowledge distillation method. 

On the other end of the spectrum, spoken questions voiced by users are also vulnerable to errors due to the ASR systems that convert them to text. How the existence of ASR noise in questions affects retrieval models is not studied yet. In this work, we address the need for evaluating passage retrieval for spoken QA. To the best of our knowledge, this is the first work in this direction. 

Since there is no available dataset for passage retrieval where questions have ASR noise, we simulate ASR noise by automatically transcribing synthetically voiced questions. We then compare the robustness of lexical and dense retrievers by evaluating them against questions with and without ASR noise. Preliminary results showed that neither lexical nor dense models are effective against questions with ASR noise leading to a significant drop in retrieval performance. We find that using data augmentation with ASR noise to train a dense retriever is a promising approach for increasing robustness against ASR noise. However, the generation of such synthetic data is time-consuming and limited to the languages/accents supported by the text-to-speech system. To this extent, we explore if typo augmentation (faster and not bound to specific accents/languages) can alleviate these limitations. Our experimental results show that typo robust dense retrievers can increase robustness against ASR noise to some extent;  however, ASR data augmentation remains a significantly more effective approach. Since users can have different local accents, we experiment with multiple accents of the same language and unveil that users' accents play an important role in retrieval performance. Finally, to study a real-world scenario with voice variation and non-native speakers voicing questions, we also build a new test set where the questions have natural ASR noise. This set consists of 700 questions voiced by human annotators.

We aim to answer the following research questions:
\textbf{RQ1} What is the impact on the performance of lexical and dense retrievers when questions have ASR errors?
\textbf{RQ2} Are typo-robust dense retrieval approaches also robust against ASR noise? How competitive are they against dense retrieval trained via data augmentation with ASR noise?
\textbf{RQ3} Do certain accents affect the effectiveness of the retriever more than others? 
\textbf{RQ4} Does natural ASR noise affect the robustness of dense retrievers more than synthetic ASR noise?

Our main contributions can be summarized as follows: (i) we provide two large-scale datasets where questions have synthetic ASR noise to facilitate research (evaluation and training of new models) on passage retrieval for spoken QA, (ii) we create a new challenging test set that contains 700 questions with natural ASR noise, (iii) we show how lexical and dense retrievers are not robust against ASR noise and propose data augmentation for robustifying the latter, and (iv) we study how performance varies with respect to different accents and synthetic vs. natural spoken questions. \footnote{\url{https://github.com/GSidiropoulos/passage_retrieval_for_spoken_qa}}
\vspace{-0.5em}
\section{Experimental Setup}

\subsection{Datasets and Evaluation Metrics}
For our experiments, we focus on two large-scale datasets, namely, MS MARCO passage ranking \cite{DBLP:conf/nips/NguyenRSGTMD16} and Natural Questions (NQ) \cite{DBLP:journals/tacl/KwiatkowskiPRCP19}. In MS MARCO the objective is to rank passages based on their relevance to a question. The questions were compiled from Bing search logs, while the underlying corpus consists of 8.8 million passages. NQ is an open-domain QA dataset where questions were sampled from Google search logs and can be answered over Wikipedia.

For MS MARCO, to measure the retrieval performance, we use the official metric MRR@10 alongside Recall (R). We report the metrics on the development set, MSMARCO (Dev),  since the ground-truths for the test set are not available to the public. Similar to previous works on NQ, we report answer recall (AR) at the top-k retrieved passages. Answer recall evaluates whether the ground-truth answer string appears among the top retrieved passages.
\vspace{-0.5em}
\subsection{Simulating ASR Noise}
\label{sec:ASR_sim}
To study the impact of speech recognition errors in passage retrieval for spoken QA, we need a large dataset of questions with ASR noise. There is no such dataset publicly available and hence in this work we build one. Following previous works \cite{DBLP:conf/interspeech/LeeWLL18, DBLP:conf/emnlp/FaisalKAA21, DBLP:conf/eacl/RavichanderDRMH21}, to simulate ASR noise, we follow a pipeline that consists of speech generation by a text-to-speech system and transcription of the generated speech by a speech-to-text system. We obtain the spoken version of the original questions via Google TTS and their transcriptions using \textit{wav2vec 2.0} \cite{DBLP:conf/nips/BaevskiZMA20}. We use English as the system language. In particular, for U.S English (en-US), the word error rate (WER) is $20.70$ and $34.26$ for NQ and MS MARCO, respectively. However, as users can have different local accents, we experiment with other English variations supported by Google TTS, such as Australian English (en-AU) and Indian English (en-IN). We report their WER scores in Table \ref{tab:wer}. Common errors in the transcribed questions include incorrect splits and phonetical spelling of relatively~rare~words~such~as~entity~mentions.

\begin{table}[]
\caption{WER for transcribed synthetic and natural spoken questions. For synthetic we report on the whole NQ (test) while for the natural on a 700-question subset of NQ (test).}
\begin{tabular}{@{}llll@{}}
\toprule
   & \multicolumn{3}{c}{Synthetic} \\
   & en-US  & en-AU  & en-IN \\ \midrule
NQ & 20.70  & 21.01  & 24.77 \\ \bottomrule
\end{tabular}

\begin{tabular}{@{}llllll@{}}
   & \multicolumn{5}{c}{Natural} \\
   & $A_1$  & $A_2$  & $A_3$  & $A_4$  & avg \\ \midrule
NQ & 38.82  & 38.16  & 44.01  & 35.28  & 39.60\\ \bottomrule
\end{tabular}
\label{tab:wer}
\end{table}
\vspace{-0.5em}
\subsection{Natural ASR Noise}
\label{sec:ASR_natural}
To simulate a natural real-world setting, we manually construct a dataset with natural ASR noise. In order to create a dataset with natural ASR noise we use the SANTLR \cite{DBLP:conf/interspeech/LiZDBM19} speech annotation toolkit. Specifically, we use SANTLR to record spoken versions of the question from four human annotators. Subsequently, we transcribe the obtained recordings using \textit{wav2vec 2.0} (similar to Section \ref{sec:ASR_sim}). We obtain audio recordings (in English) for spoken versions of 700 questions from the NQ dataset, voiced by four human annotators. The annotators were instructed to (i) read the prompt question, (ii) ensure that they can pronounce every word appearing in the question or move to the next one, and (iii) finally record. The annotators consisted of a French female ($A_1$), a Greek male ($A_2$), an Indian male ($A_3$), and a Russian female ($A_4$), with the first three voicing 200 unique questions each and the last voicing 100 unique questions. All annotators are using English in their everyday life. We use a mixture of accents originating from non-native English speakers to resemble a real-world scenario. Voice assistants do not support the majority of the world's languages \cite{DBLP:journals/corr/abs-2204-08582}. Therefore, many users have to voice their questions in a language different from their native one. WER scores can be found in Table \ref{tab:wer}. Similar to synthetic noise, the most common errors include incorrect splits and the phonetical spelling of entity mentions. However, these errors are significantly more prominent in the case of natural noise.
\vspace{-0.5em}
\subsection{Models}

\label{sec:models}
\textbf{BM25} is a standard retrieval model based on best match; there is lexical overlap between the query and every retrieved passage. We use the Anserini IR toolkit \cite{DBLP:conf/sigir/Yang0L17} to compute BM25 scores.

\textbf{Dense Retriever (DR)} \cite{DBLP:conf/emnlp/KarpukhinOMLWEC20} is a dual-encoder BERT-based model used for scoring question-passage pairs. Given a question $q$, a positive (i.e., relevant) passage $p^+$ and a set of negatives (i.e., irrelevant) passages $\{p_1^-, p_2^-, \dots, p_n^-\}$, the model learns to rank the positive question-passage pair higher than the negative ones. The two seperate encoders of the model are fine-tuned via the minimization of the softmax cross-entropy:

\begin{equation}
\mathcal{L}_{CE} = -\log \frac{e^{s(q,p^+)}}{e^{s(q,p^+)} + \sum_{p^-}e^{s(q,p^-)}}.
\label{eq:ce}
\end{equation}

During inference time, the similarity of a question-passage pair is calculated as the inner product of the respective question embedding and passage embedding. In detail, the whole corpus is encoded into an index of passage vectors offline, and the retrieval with respect to a question is implemented with efficient maximum inner product search ~\cite{DBLP:journals/tbd/JohnsonDJ21} over the index. We follow the dual-encoder architecture compared to a cross-encoder one (that jointly encodes question and passage) due to its high efficiency as a first-stage ranker in large-scale settings. While the latter can achieve higher performance, the former makes the whole corpus indexable.

\textbf{Dense retriever with data augmentation (DR+Data augm.)} builds on the standard practice for improving the robustness of neural models by augmenting the training data with noisy data. We explore two cases of data augmentation, namely, augmentation with synthetic keyboard noise and augmentation with synthetic ASR noise. For the former, we augment each question on the training set with keyboard noise following the approach presented in \cite{ DBLP:conf/sigir/SidiropoulosK22}, while for the latter, we augment with ASR noise as shown in Section \ref{sec:ASR_sim}. In contrast with ASR noise, keyboard noise will rarely alter the original word into a different correctly spelled word.
For example, the question "who is the owner of reading football club" can be transformed to "who is the owner of retting football club" if augmented with ASR noise and to "who is the ownrr of reading football club" if augmented with keyboard noise; ``retting" is a correctly spelled word, while ``ownrr" is not.

\textbf{Dense retriever with characterBERT and self-teaching (DR+characterBERT+ST)} \cite{DBLP:journals/corr/abs-2204-00716} is the current state-of-the-art dense retrieval approach for dealing with typos. It builds on DR by altering the backbone BERT encoder with CharacterBERT and further uses an effective training method that distills knowledge from questions without typos into the questions with typos, known as self-teaching. Specifically, the goal of the latter is to minimize the difference between the similarity score distribution from the question with the typo and the score distribution from the question without the typo. This is achieved by minimizing the KL-divergence:

\begin{eqnarray}
\mathcal{L}_{KL} =  \tilde{s}(q',p) \cdot \log \frac{\tilde{s}(q',p)}{ \tilde{s}(q,p)},
\label{eq:kl}
\end{eqnarray}

where $q'$ represents the typoed question and $\tilde{s}$ the softmax normalized similarity score. The final loss is the sum of the $\mathcal{L}_{KL}$ (Equation \ref{eq:kl}) and $\mathcal{L}_{CE}$ (Equation \ref{eq:ce}) losses.
\vspace{-0.5em}
\subsection{Implementation Details}
\label{sec:details}
The DR model we use in our experiments is trained using the in-batch negative setting described in \cite{DBLP:conf/emnlp/KarpukhinOMLWEC20}. The question and passage BERT encoders are trained using Adam with a learning rate of $2e$-$5$ and linear scheduling with warm-up rate of $0.1$ for (i) $40$  epochs with a batch size of $64$ for the case of NQ and (ii) $10$ epochs with a batch size of $84$ for MS MARCO. Moreover, we use the same hyper-parameters when training DR with data augmentation. For DR+characterBERT+ST, we use the pre-trained model as provided by the authors of \cite{DBLP:journals/corr/abs-2204-00716}. The audio input is sampled at 16Khz to be compatible with the transcription model we use (\textit{wav2vec}).
\section{Results}

\begin{table*}[]
\caption{Retrieval results for the settings of (i) clean questions (Original) and (ii) questions with synthetic ASR noise (ASR); synthetic voice with a U.S. English accent. Statistical significance difference with paired t-test $(p < 0.05)$ BM25=b, DR=d; DR with typo data augm.=t; DR+CharBERT+ST=c. Note that we use the pretrained  DR+CharBERT+ST from the original paper.}
\resizebox{\textwidth}{!}{%
\begin{tabular}{@{}l|l|llllll|llllll@{}}
\toprule
\multirow{3}{*}{} &
  \multirow{3}{*}{Noise} &
  \multicolumn{6}{c|}{NQ (Test)} &
  \multicolumn{6}{c}{MS MARCO (Dev)} \\ \cmidrule(l){3-14} 
     &       & \multicolumn{3}{c}{Original} & \multicolumn{3}{c|}{ASR} & \multicolumn{3}{c}{Original} & \multicolumn{3}{c}{ASR} \\
     &       & AR@5    & AR@20   & AR@100   & AR@5   & AR@20 & AR@100 & R@50    & R@1000   & MRR@10  & R@50  & R@1000 & MRR@10 \\ \midrule
BM25 & -     & 40.94   & 57.81   & 70.83    & 23.32  & 36.98 & 52.49  & 59.11   & 85.61    & 18.67   & 24.71 & 45.34  & 6.97   \\
DR   & -     & 66.26   & 77.75   & 85.26    & 41.91  & 54.65 & 67.03  & 74.58   & 94.19    & 28.69   & 35.31 & 56.83  & 12.13  \\
DR+Data augm.   & Typos & 67.47   & 78.75   & 85.40    & 46.75  & 60.72 & 71.55  & 75.17   & 94.54    & 29.10   & 46.75 & 64.57  & 13.02  \\
\begin{tabular}[c]{@{}l@{}}DR+CharBERT+\\+ST\cite{DBLP:journals/corr/abs-2204-00716}\end{tabular}   & Typos & -   & -   & -    & -  & - & -  & 77.55   & 94.95    & 32.51   & 45.45 & 68.20  & 16.35  \\
DR+Data augm.   & ASR   & 66.67   & 78.00   & 85.45    & \textbf{54.84}$^{bdtc}$  & \textbf{67.89}$^{bdtc}$ & \textbf{78.50}$^{bdtc}$  & 73.47   & 93.96    & 29.14   & \textbf{54.48}$^{bdtc}$ & \textbf{81.25}$^{bdtc}$  & \textbf{18.43}$^{bdtc}$  \\ \bottomrule
\end{tabular}%
}
\label{tab:original_vs_synthetic}
\end{table*}

In this section, we present our experimental results
that answer our research questions. To answer \textbf{RQ1}, we compare the retrieval performance of a lexical retriever (BM25) and a dense retriever (DR) for the settings of clean questions and questions with synthetic ASR noise. As we can see from the first two rows in Table \ref{tab:original_vs_synthetic},  DR significantly outperforms BM25 across both settings (original and ASR). On the other hand, when questions have ASR noise, there is a dramatic drop in performance for both BM25 (MRR@10 drops $62.66\%$ and AR@5 $43.03\%$) and DR (MRR@10 drops $57.72\%$ and AR@5 $36.74\%$). This drop indicates the lack of robustness of lexical and dense models against ASR noise. 

Data augmentation with typos and self-teaching for knowledge distillation from questions without typos into the questions with typos are two training schemes for robustifying DR. For \textbf{RQ2}, we examine how these two perform compared to DR with standard training on clean questions on the ASR noise scenario. For rows using ``Typos" as noise in Table \ref{tab:original_vs_synthetic}, we notice that the models can increase robustness against ASR noise, to some extent, even though the typos are not originating from the same distribution as the ASR noise during inference. That said, as was expected, DR holds the best results when augmenting the training set with ASR noise.

Furthermore, we investigate how the retrieval performance varies depending on the number of available questions with ASR noise used to augment the training set. Table \ref{tab:data_aug} shows the results. We observe a significant increase in performance even in the low data regime, with 400 noisy questions. Intuitively, there is consistent improvement in performance as the additional data increase.

\begin{table}[]
\caption{Retrieval results for DR trained with ASR data augmentation (questions augmented with U.S. English synthetic ASR noise).}
\begin{tabular}{llll}
\hline
\begin{tabular}[c]{@{}l@{}}\# Additional training\\ questions with ASR noise\end{tabular} & AR@5 & AR@20 & AR@100 \\ \hline
40K & 54.84 & 67.89 & 78.50 \\
4K  & 48.69 & 61.96 & 73.35 \\
400 & 44.29 & 57.28 & 69.27 \\
0   & 41.91 & 54.65 & 67.03 \\ \hline
\end{tabular}%
\label{tab:data_aug}
\end{table}

At this point, we have seen that DR with ASR data augmentation is an effective approach for dealing with spoken questions. For \textbf{RQ3}, we want to study the impact of different English accents during inference. To do so, we use the DR model which is augmented with synthetic ASR noise from U.S English and test it against different spoken accents such as Australian English (en-AU) and Indian English (en-IN). The results in Table \ref{tab:synthetic_accents} show that different accents have different impacts on performance. Specifically, we observe a small drop in performance for Australian English while the drop is more prominent for Indian English. This is strongly related to the fact that U.S English is phonetically more similar to Australian English compared to Indian English.

We underline that using synthetic ASR noise in the respective English variation for augmenting the training set could help boost performance. However, this is not a viable solution if we take into account that (i) such an approach would require training a new system for every new variation and to the extreme for each individual, and (ii) there is a limitation in the available synthetic voice accents. 

\begin{table}[]
\caption{Retrieval results for DR trained with ASR data augmentation (augmented with U.S. English synthetic ASR noise) and tested against synthetic ASR for various accents.}
\begin{tabular}{@{}lllll@{}}
\toprule
Data augm. & test  & AR@5  & AR@20 & AR@100 \\ \midrule
en-US      & en-US & 54.84 & 67.89 & 78.50  \\
en-US      & en-AU & 53.54 & 66.70 & 76.92  \\
en-US      & en-IN & 46.37 & 60.27 & 71.85  \\ \bottomrule
\end{tabular}%

\label{tab:synthetic_accents}
\end{table}

In a real-world scenario, alongside voice variation, accents can greatly vary depending on the origin of the user; since non-native English speakers are voicing their questions in English when interacting with voice assistants. To showcase the real-world utility of dense retrieval (\textbf{RQ4}), we evaluate DR and its data augmented variations on our natural ASR noise dataset (see Section \ref{sec:ASR_natural}). As we can see in Table \ref{tab:synthetic_vs_natural}, even though both synthetic and natural ASR noise decreases the retrieval performance, natural ASR noise appears to be a significantly more challenging setting. Despite the fact that the distribution of synthetically generated ASR noise and keyboard noise differs from that of natural ASR, we notice that DR combined with data augmentation holds better results than DR alone. We find that DR with synthetic ASR data augmentation outperforms its typo counterpart. Unfortunately, synthetic ASR and typo noise are inefficient while they do not generalize well to natural ASR noise.

\begin{table}[]
\caption{Retrieval results for DR against questions with ASR noise from (i) synthetic voice with a U.S. English accent (Synthetic), and (ii) natural voice with a mixture of French, Greek, Indian and Russian accents (Natural). We use the same subset of 700 questions to ensure a fair comparison.}
\resizebox{\columnwidth}{!}{%
\begin{tabular}{@{}l|lll|llll@{}}
\toprule
\multirow{2}{*}{\begin{tabular}[c]{@{}l@{}}Data \\ augm.\end{tabular}} & \multicolumn{3}{c|}{ASR Synthetic} & \multicolumn{3}{c}{ASR Natural} \\
      & AR@5  & AR@20 & AR@100 & AR@5  & AR@20 & AR@100 \\ \midrule
-     & 38.88 & 53.65 & 65.85  & 16.64 & 28.55 & 42.46  \\
Typos & 46.91 & 59.82 & 71.87  & 24.82 & 35.86 & 50.50  \\
ASR   & 55.38 & 67.43 & 77.76  & 29.98 & 42.32 & 55.66  \\ \bottomrule
\end{tabular}%
}
\label{tab:synthetic_vs_natural}
\end{table}

\section{Conclusions}
In this work, we study the impact of speech recognition errors in passage retrieval for spoken QA. We showcase that the effectiveness of lexical and dense retrievers drops  dramatically when dealing with transcribed spoken questions. Moreover, we explore how typo-robust dense retrieval approaches perform against questions with ASR noise. Even though they can increase robustness compared to standard training on clean questions only, dense retrieval trained via data augmentation with ASR noise is a more effective approach. Finally, we compare the effect of synthetic vs. natural ASR noise and find that the latter is a significantly more challenging setting. Unfortunately, data augmentation with synthetic ASR noise does not generalize well to the natural ASR scenario. 
For future work, we plan to build on our insights and develop more sophisticated approaches for robustifying dense retrievers against ASR noise. Also, we aim to investigate ways to produce noise that can closely resemble natural ASR noise.

\partitle{Acknowledgments.}
This research was supported by
the NWO Innovational Research Incentives Scheme Vidi (016.Vidi.189.039),
the NWO Smart Culture - Big Data / Digital Humanities (314-99-301),
the H2020-EU.3.4. - SOCIETAL CHALLENGES - Smart, Green And Integrated Transport (814961).
All content represents the opinion of the authors, which is not necessarily shared or endorsed by their respective employers and/or sponsors.

\clearpage
\newpage
\bibliographystyle{ACM-Reference-Format}
\balance
\bibliography{main}

\end{document}